\documentclass[runningheads]{llncs}
\usepackage[T1]{fontenc}
\usepackage{subcaption}
\usepackage{graphicx}
\usepackage{booktabs}
\usepackage{amsmath}
\usepackage{float}
\usepackage{hyperref}
\usepackage{color}

\title{Semantic Similarity in Radiology Reports via LLMs and NER}
\date{}

\institute{}
\begin{document}
\author{Beth Pearson\inst{1} \and
Ahmed Adnan\inst{2} \and
Zahraa S. Abdallah\inst{1}}
%

%
\institute{
University of Bristol, Bristol, UK \\
\email{\{beth.pearson, zahraa.abdallah\}@bristol.ac.uk}
\and
Rosenfield Health Tech Ltd, Cardiff, UK \\
\email{ahmed.adnan@rosenfieldhealth.com}
}
\maketitle
\begin{abstract}
    Radiology report evaluation is a crucial part of radiologists' training and plays a key role in ensuring diagnostic accuracy. As part of the standard reporting workflow, a junior radiologist typically prepares a preliminary report, which is then reviewed and edited by a senior radiologist to produce the final report. Identifying semantic differences between preliminary and final reports is essential for junior doctors, both as a training tool and to help uncover gaps in clinical knowledge. While AI in radiology is a rapidly growing field, the application of large language models (LLMs) remains challenging due to the need for specialised domain knowledge. In this paper, we explore the ability of LLMs to provide explainable and accurate comparisons of reports in the radiology domain. 
    We begin by comparing the performance of several LLMs in comparing radiology reports. We then assess a more traditional approach based on Named-Entity-Recognition (NER). However, both approaches exhibit limitations in delivering accurate feedback on semantic similarity. To address this, we propose Llama-EntScore, a semantic similarity scoring method using a combination of Llama 3.1 and NER with tunable weights to emphasise or de-emphasise specific types of differences. Our approach generates a quantitative similarity score for tracking progress and also gives an interpretation of the score that aims to offer valuable guidance in reviewing and refining their reporting. We find our method achieves 67\% exact-match accuracy and 93\% accuracy within ± 1 when compared to radiologist-provided ground truth scores — outperforming both LLMs and NER used independently. Code is available at: \href{https://github.com/otmive/llama_reports}{github.com/otmive/llama\_reports}
    
\end{abstract}

\section{Introduction}
Radiology reports are important medical documents that capture the key findings from imaging modalities such as magnetic resonance imaging (MRI) scans, computed tomography (CT) scans and ultrasounds. In clinical practice, a preliminary report is drafted by a junior radiologist and subsequently reviewed by a senior radiologist who makes edits and produces the final report. Identifying semantic differences between preliminary and final radiology reports is essential for supporting the training of junior radiologists and improving the overall reporting process. These differences often reflect important corrections, clarifications, or additions made by senior radiologists and can reveal specific areas where clinical understanding can be strengthened. However, manually reviewing large numbers of reports to extract such insights is time-consuming and difficult to scale. Automating this process enables systematic feedback, supports targeted learning, and offers the potential for AI tools to assist in evaluating report quality and consistency across the workflow.~\cite{sharpe2012radiology,harari2016role,kalaria2016comparison,alam2020comparative}. While radiologists find these automated systems beneficial, particularly those with a visualisation of differences~\cite{sharpe2012radiology,kalaria2016comparison}, the comparisons often fall short of expectations. Most rely on basic text comparisons, which fail to capture the clinical relevance or semantic nature of the differences, and do little to highlight broader trends or educational value. Radiology reports contain dense medical jargon and highly specialised terminology, making them particularly challenging to interpret, especially for methods not specifically designed to handle such complexity. More nuanced NLP techniques, such as NER using models trained on biomedical corpora have been explored in previous work ~\cite{picha2024semantic,jain2021radgraph,kalaria2016comparison}. However, these approaches often struggle to match the depth and nuance of feedback typically provided by experienced radiologists.

This paper investigates the potential of large language models (LLMs) to automate the semantic comparison of radiologist reports. We then propose a hybrid approach that combines LLMs with Named-Entity-Recognition (NER) to generate both a numerical similarity score and a qualitative interpretation. This dual output is designed to support feedback for junior radiologists and help identify common trends in report reviews. The inclusion of NER helps to ground the score in clinically relevant content by enabling a direct comparison between technical entities contained in both reports. Our use case focuses on providing structured, scalable feedback to junior doctors, allowing them to identify recurring gaps in their preliminary reports and track improvements over time.
Due to the sensitive nature of radiology report data, we focus on open-source models which can be deployed on local servers to maintain patient confidentiality. 
We evaluate four LLMs on their ability to assess the semantic similarity of radiology reports and review the quality of their outputs. Building on these insights, we introduce Llama-EntScore - a novel method for comparing preliminary and final radiology reports using Llama 3.1 and NER to give a numerical score and interpretative feedback. While we use a base model LLM, the NER model has been pre-trained on biomedical data, enhancing the method's domain-specific relevance. The remainder of this paper is organised as follows: Section 2 reviews related work, Section 3 presents the proposed methods, Section 4 explains the results, and the paper concludes in Section 5.

\section{Related Work}
LLMs have been applied for a range of tasks in the radiology domain, including report generation~\cite{wang2023r2gengpt}, information extraction~\cite{le2024performance} and report summarisation~\cite{hu2024current,liu2023radiology}. Several studies have explored the use of LLMs to compare AI-generated reports with an expert-written ground truth~\cite{wang2024llm,xu2024reasoning,zhu2024leveraging}, though comparing two human-written reports remains relatively unexplored. Zhu et al. ~\cite{zhu2024leveraging} propose a method for comparing AI-generated reports with a ground truth using GPT-3.5 and GPT-4 with in-context learning.
Bala et al.~\cite{bala2024enhancing} use prompt-tuning with GPT-4 to identify missed diagnoses in preliminary radiology reports. Voinea et al. use Llama 3 to summarise MRI and CT radiology reports~\cite{voinea2024gpt}.  Doshi et al. use LLMs to simplify the impressions section of radiology reports, aiming to reduce medical jargon and improve readability for patients~\cite{doshi2024quantitative}.

Several specialised models have also been developed for specific radiology tasks. CheXbert~\cite{smit2020chexbert} is a model for labelling chest X-ray reports using a biomedically pre-trained version of BERT~\cite{devlin2019bert}. Tasks include marking specific diagnoses as positive, negative, or uncertain. However, this method is limited by its focus on chest X-rays only. RadCLiQ is a metric for evaluating cross-over entities between AI-generated and radiologist reports~\cite{yu2023evaluating}. RadCLiQ combines four existing metrics, including RadGraph~\cite{jain2021radgraph} - a language model trained on chest X-ray reports. 

Recent work has explored the combination of LLMs and NER to handle tasks in the biomedical domain. Hu et al. ~\cite{hu2024improving} use GPT models to extract clinical entities from texts and develop task-specific prompts to refine their results. Ghali et al. ~\cite{ghali2024gamedx} use in-context learning to leverage open-source LLMs for medical data extraction. Picha et al. ~\cite{picha2024semantic} create a specific cosine similarity metric for chest X-ray reports using NER to extract key entities for calculating similarity. Zhang et al. ~\cite{zhang2024constructing} fine-tune an LLM for generating radiology report impressions from a given list of findings. 

While prior work has largely focused on comparison involving AI-generated content or limited modalities such as chest X-rays, there is limited exploration of how LLMs can be used to compare two human-written radiology reports across broader imaging modalities. This paper addresses that gap by combining LLMs with NER to assess semantic similarities between preliminary and final radiology reports, offering feedback for junior radiologists during training.

\section{Methodology}

We propose a hybrid approach that combines NER and LLMs to compute both a semantic similarity score and an interpretable explanation of the differences between preliminary and final radiology reports. The goal is to produce feedback that aligns more closely with the nuanced understanding of a radiology expert. We begin by describing the NER component of the method.

\subsection{Entity-Based Scoring via NER}
\label{sec:ner}
NER is a widespread natural language processing (NLP) technique that is used to extract key information or clinical entities from text. In our approach, we extract entities from both the preliminary and final radiology reports and compute a cosine similarity score to assess semantic overlap based on the MCSE score from Picha et al.~\cite{picha2024semantic}. 


We begin by identifying entities that exactly match between the two reports. These are classified as matched entities. For each unmatched entity in the final report, we compute similarity scores with all unmatched entities in the preliminary report and assign the highest similarity score to the corresponding entity. The overall similarity score is then computed as:
\begin{equation}
\label{eq:cosine}
Score_{NER}= \frac{C + \sum_{i=1}^{N} \max(S_j)}{T}, \quad j = 1, ..., M
\end{equation}

where $C$ is the number of matched entities, $N$ and 
$M$ are the number of unmatched entities in the final and preliminary reports, respectively, and $T$ is the total number of entities in the final report.

While the NER-based similarity score offers a structured way to quantify differences between reports, it is inherently limited to surface-level comparisons. It cannot account for how an entity is used — whether a finding is negated, modified in severity, or described differently. These subtleties are often critical in clinical interpretation. To address this, we extend our method by incorporating an LLM that evaluates the contextual meaning of each shared entity, allowing us to distinguish not just presence but semantic intent.

\subsection{Context-Aware Scoring with NER and LLMs}
\label{sec:context-aware}

To address these limitations, we extend the method by introducing a context-sensitive scoring mechanism that incorporates the semantic use of entities, as judged by an LLM. We refer to this enhanced metric as the \textbf{Entity-Based Semantic Agreement Score (ESAS)}.

After extracting named entities from both reports, we first identify entities that are shared between the two. For each shared entity, we assess whether it is used in the same clinical context by prompting \textbf{LLaMA 3.1} with the following query:

\begin{quote}
\textit{Can you say whether the entity: '\{entity\}' is used in the same context or different context in these two texts? \\
Text1: '\{report1\}' \\
Text2: '\{report2\}' \\
Please reply with a single-word answer: either 'same' or 'different'.}
\end{quote}

\noindent
Entities are then categorised based on both their presence and LLM response:

\begin{itemize}
    \item \textbf{Matched}: Shared entities judged by the LLM to be used in the same context.
    \item \textbf{Mismatched}: Shared entities judged by the LLM to be used differently (e.g., differing severity or negation).
    \item \textbf{Missing}: Entities present only in the final report.
    \item \textbf{Surplus}: Entities present only in the preliminary report.
\end{itemize}

\noindent
We compute the ESAS as follows:

\begin{equation}
\text{ESAS} = \frac{N_{\text{match}}}
{N_{\text{match}} + \sum\limits_{c \in \{ \text{mismatch}, \text{missing}, \text{surplus} \}} W_c \cdot N_c}
\label{eq:esas_generalised}
\end{equation}

\noindent
Where
 $N_{\text{match}}$ is the number of contextually matched entities, $N_c$ is the number of entities in category $c$, where $c \in \{ \text{mismatch}, \text{missing}, \text{surplus} \}$, and $W_c$ is the penalty weight assigned to category $c$ to reflect its clinical or educational impact. 

The four categories were selected as they provide a robust framework for comparing differences that remain relevant across all scan types, and they align with the criteria radiologists typically consider when manually reviewing reports.
The weighting values are designed to reflect the relative clinical and educational significance of each type of discrepancy. 
The weight $W_{\text{missing}}$ corresponds to missing clinically important content in the preliminary report, which is penalised most heavily, as such omissions may indicate critical gaps in knowledge or pose risks to diagnostic accuracy. 
In contrast, $W_{\text{mismatch}}$ accounts for contextual mismatches—such as incorrect severity, anatomical location, or negation—which are penalised slightly less but still reflect meaningful misunderstandings. 
Finally, $W_{\text{surplus}}$ represents surplus entities, i.e., content present in the preliminary report but not retained in the final version. These are penalised the least, as they often reflect over-reporting or stylistic variation rather than substantive clinical error. 

The weighting scheme is tunable and can be adapted for different training objectives—allowing for stricter evaluation in early training phases or more lenient interpretation when encouraging thorough documentation. In this way, ESAS provides a flexible, interpretable, and clinically grounded metric for assessing semantic similarity between radiology reports.

\subsection{LLM-Based Interpretation of Similarity Scores}

Following the calculation of the ESAS, we prompt the LLM to provide a qualitative explanation of the result. We observed that anchoring the prompt with a numerical score helps reduce hallucinations and guides the model toward more relevant, technical interpretations. The prompt used is:

\begin{quote}
\textit{These two reports have a similarity score of \{score\}. Report 1 is the final report, and Report 2 is the preliminary report. \\
Can you give a reason for the similarity score? Focus on technical details rather than structure or style. \\
Report 1: \{report1\} \\
Report 2: \{report2\}}
\end{quote}

This hybrid approach combines the structured, domain-specific strengths of NER with the interpretive capabilities of LLMs, resulting in both a quantitative similarity score and a human-readable explanation. To further support interpretability, we also provide a visualisation of the identified entities and their classification into \textit{matched}, \textit{mismatched}, \textit{missing}, or \textit{surplus}. An example of this visualisation is shown in Figure~\ref{fig:entities}.




\section{Results}
\subsection{Dataset}

Our dataset consists of 115 anonymised pairs of radiology reports, each comprising a preliminary report written by a junior radiologist and a corresponding final report reviewed by a senior radiologist. The dataset includes 56 MRI scans, 48 CT scans, 2 ultrasound scans, and 9 reports for which the imaging modality is unknown. Each report typically includes two sections: a \textit{Findings} section, which details the observed anatomical and pathological features, and an \textit{Impression} section, which summarises and prioritises the clinically significant findings. While our dataset is relatively small, it includes a variety of scan types from two different sources, one in the UK and one in Qatar, which supports the generalisability of our method to diverse clinical scenarios.
We evaluate both the numerical similarity scores generated by our method and the quality of the corresponding LLM-generated interpretations. All data used in this study comes from a private, anonymised clinical dataset. Experiments are carried out in a Linux environment using an RTX 2080 GPU.
\subsection{Evaluating LLMs}
\label{sec:llms}

We evaluated four LLMs on their ability to compare radiology report pairs: \textbf{LLaMA-3.1-Instruct-8B}~\cite{dubey2024llama}, \textbf{LLaMA-2-Chat-7B}~\cite{touvron2023llama}, \textbf{Mistral-Instruct-7B}~\cite{jiang2023mistral}, and \textbf{BioMistral-7B}~\cite{labrak2024biomistral}. Our goal is to assess both the numerical similarity scores and the qualitative explanations generated by these models.

To test sensitivity to subtle semantic changes, we designed a small-scale experiment comparing (1) pairs of identical reports and (2) report pairs differing by a single negation. For example, a report stating “back muscle spasm” in the final version might have “no back muscle spasm” in the preliminary. These minimal negation changes are clinically significant but lexically similar, making them useful for evaluating contextual understanding.

To generate these modified reports, we used LLaMA-3.1 to create near-identical variations of existing reports with only one negation difference. The prompt used is:

\begin{quote}
\textit{Generate a report identical to this one but with one negation change, e.g., “broken arm” becomes “no broken arm”. Please only make one change from the original report.}

\texttt{Report: \{report\_1\}}

\textit{Please only output the report, no other text.}
\end{quote}

\begin{figure}[htbp]
    \centering
    \includegraphics[width=0.7\linewidth]{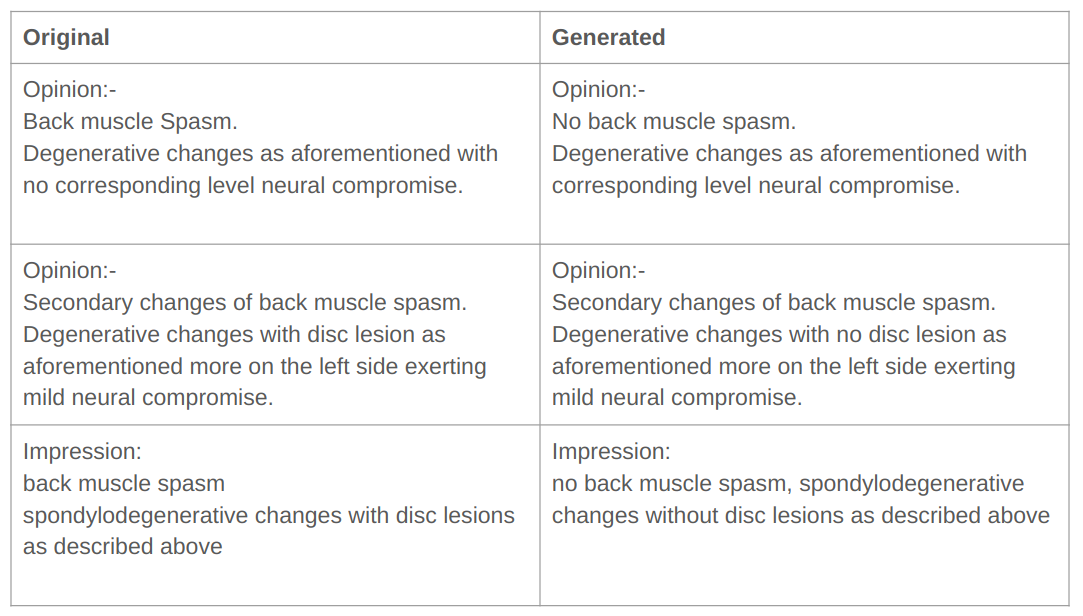}
    \caption{Examples of small negation changes in report impression sections.}
    \label{fig:small_neg}
\end{figure}

Examples of generated pairs are shown in Figure~\ref{fig:small_neg}.

Each LLM was prompted to compare the two reports using the following format:

\begin{quote}
\textit{Please provide a similarity score out of 10 for these two reports. Focus on technical content rather than style or phrasing.}

\texttt{Score: <score>, Reasoning: <reasoning>}

\texttt{Report 1: \{report\_1\}, Report 2: \{report\_2\}}
\end{quote}

Table~\ref{tab:llm_comp} shows the average similarity scores across the identical pairs and the small negation pairs.

\begin{table}[htbp]
  \begin{center}
    \resizebox{0.6\textwidth}{!}{
    \begin{tabular}{lcc}
    \toprule
    \textbf{Model} & \textbf{Identical Report Score} & \textbf{Small Negation Score} \\
    \midrule
    Mistral         & 9.3 & 8.9 \\
    BioMistral      & 8.1 & 7.9 \\
    LLaMA 2         & 8.0 & 7.8 \\
    LLaMA 3.1       & \textbf{9.7} & 8.7 \\
    \bottomrule
    \end{tabular}}
    \caption{Average similarity scores produced by each LLM for identical and small-negation report pairs.}
    \label{tab:llm_comp}
  \end{center}
\end{table}

In addition to numerical scores, we analysed the textual explanations generated by each model which provide a justification for the similarity score based on the content in each report. We observed that \textbf{LLaMA 2} and \textbf{BioMistral} were more prone to hallucinations—describing differences that were not present in the input reports. While \textbf{Mistral} and \textbf{LLaMA 3.1} also occasionally produced hallucinations, their reasoning was generally more accurate and aligned with the report content.

Overall, we found that \textbf{LLaMA 3.1} generated the most detailed and contextually faithful outputs, both in terms of reasoning and numerical score reliability. As such, we selected LLaMA 3.1 as the base LLM for our hybrid Llama-EntScore method.

Despite its superior performance, LLaMA 3.1 still tended to assign overly high similarity scores to report pairs with clinically important differences—particularly in the small-negation cases. For instance, a change from “back muscle spasm” to “no back muscle spasm” would still receive a score of 8 or higher, even though the implication for diagnosis is substantial. This motivated the development of a more structured, entity-based similarity method to provide more granular and clinically grounded scoring.

\subsection{Llama-EntScore Evaluation}
We evaluate four similarity scoring methods for radiology report comparison: (1) a word-for-word overlap metric, (2) \textit{LLaMA 3.1} alone, (3) cosine similarity based on NER, and (4) our proposed hybrid method, \textit{Llama-EntScore}. The word-for-word metric computes the proportion of words in the final report that also appear in the preliminary report. NER is performed using SciSpacy~\cite{neumann2019scispacy} to extract clinically relevant terms from both reports. We choose the large core model as we found it to retrieve the highest number of entities for our report data. The implementation details of the LLaMA-based and cosine similarity approaches are described in Sections~\ref{sec:llms} and~\ref{sec:ner}, respectively. Our hybrid method, described in Section~\ref{sec:context-aware}, combines the structured output of NER with the contextual understanding of a language model to produce a more interpretable and clinically aligned similarity score. For our experiments, the default weights in the Entity-Based Semantic Agreement Score (ESAS) are set as follows: $W_{\text{missing}} = 2$, $W_{\text{mismatch}} = 1.5$, and $W_{\text{surplus}} = 1$, reflecting their relative clinical significance. We made small adjustments to these weights and chose the values based on heuristic reasoning and alignment to ground-truth scores.

\begin{figure}[htbp]
\centering
\begin{subfigure}{.3\textwidth}
  \centering
  \includegraphics[width=\linewidth]{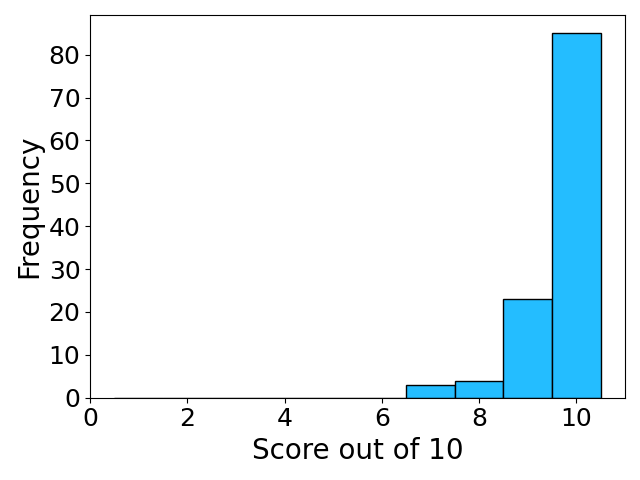}
  \caption{Distribution of Word-For-Word scores}
  \label{wfw_hist}
\end{subfigure}%
\hspace{0.03\textwidth}
\begin{subfigure}{.3\textwidth}
  \centering
  \includegraphics[width=\linewidth]{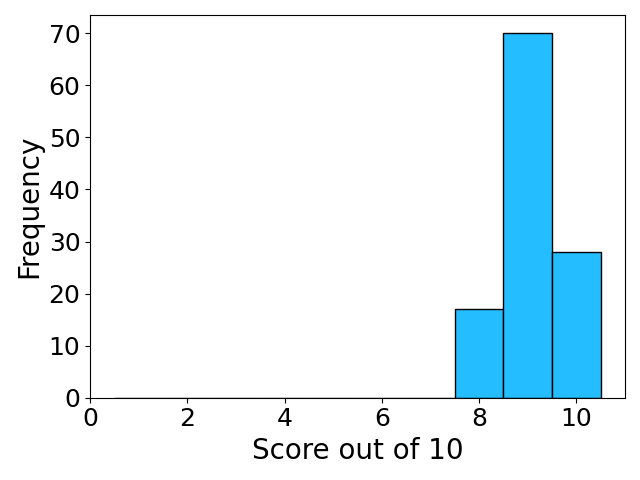}
  \caption{Distribution of Llama only scores}
  \label{llama_hist}
\end{subfigure}%
\hspace{0.03\textwidth}
\begin{subfigure}{.3\textwidth}
  \centering
  \includegraphics[width=\linewidth]{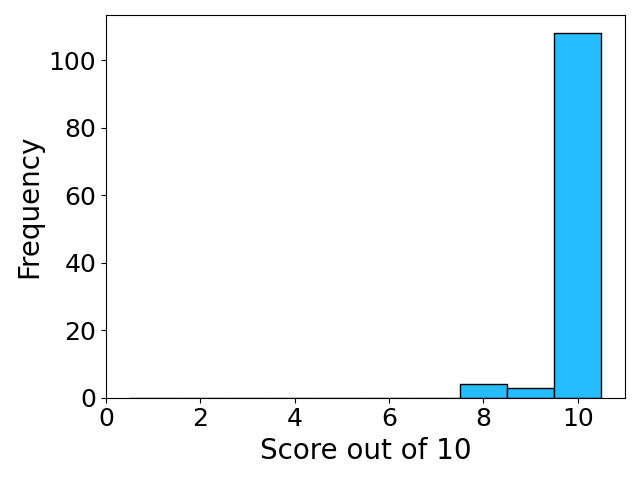}
  \caption{Distribution of NER Cosine scores }
  \label{cosine_hist}
\end{subfigure}
\vspace{0.05\textwidth}
\begin{subfigure}{.3\textwidth}
  \centering
  \includegraphics[width=\linewidth]{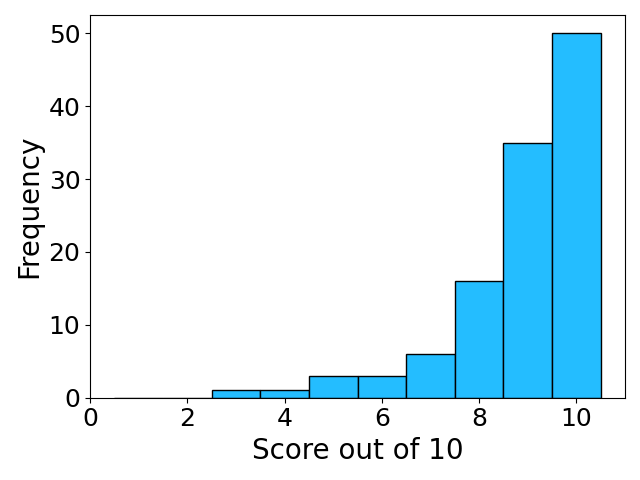}
  \caption{Distribution of Llama-EntScore}
  \label{our_hist}
\end{subfigure}%
\hspace{0.03\textwidth}
\begin{subfigure}{.3\textwidth}
  \centering
  \includegraphics[width=\linewidth]{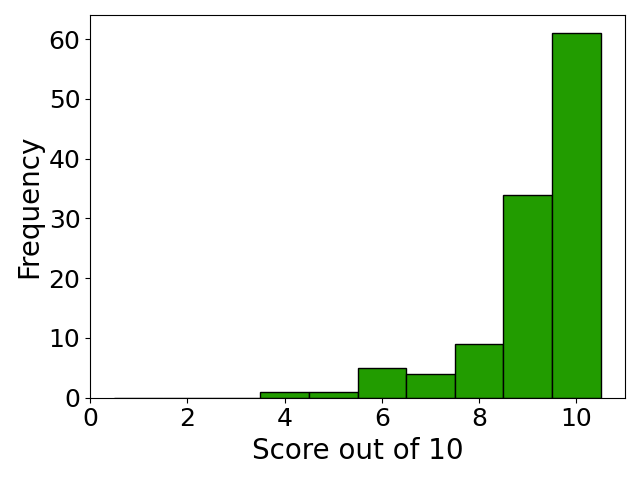} 
  \caption{Distribution of ground truth scores}
  \label{fifth_hist}
\end{subfigure}
\caption{Distribution of similarity scores for all methods. Llama-EntScore most closely follows the shape and spread of expert-annotated ground truth scores.}
    \label{fig:score_distributions}
\end{figure}

 Figure \ref{fig:score_distributions} shows the score distributions produced by each method, and the ground truth score distributions. We compare these distributions, providing insight into how closely each method aligns with the ground truth scores assigned by radiology experts.  Among the four approaches, Llama-EntScore demonstrates the closest match to expert annotations, with a wider spread of scores and better differentiation in mid-range values. Other methods tend to over-predict high similarity: the word-for-word and cosine methods cluster around perfect matches, while Llama 3.1 alone centres around 9 regardless of subtle changes.

Quantitative results are summarised in Table~\ref{tab:results}. Accuracy and related metrics are calculated by rounding predictions and comparing them to the integer-rounded ground truth scores. Therefore,  a ground truth score of, for example, 9.3 and 9.2 are treated as a match. Llama-EntScore outperforms all other methods across every metric, with a 10\% gain in strict accuracy over the next-best method. Its precision and recall also indicate more reliable alignment with expert judgment

\begin{figure}[h!]
  \centering

  \begin{subfigure}[b]{0.45\textwidth}
    \centering
    \includegraphics[width=0.9\linewidth]{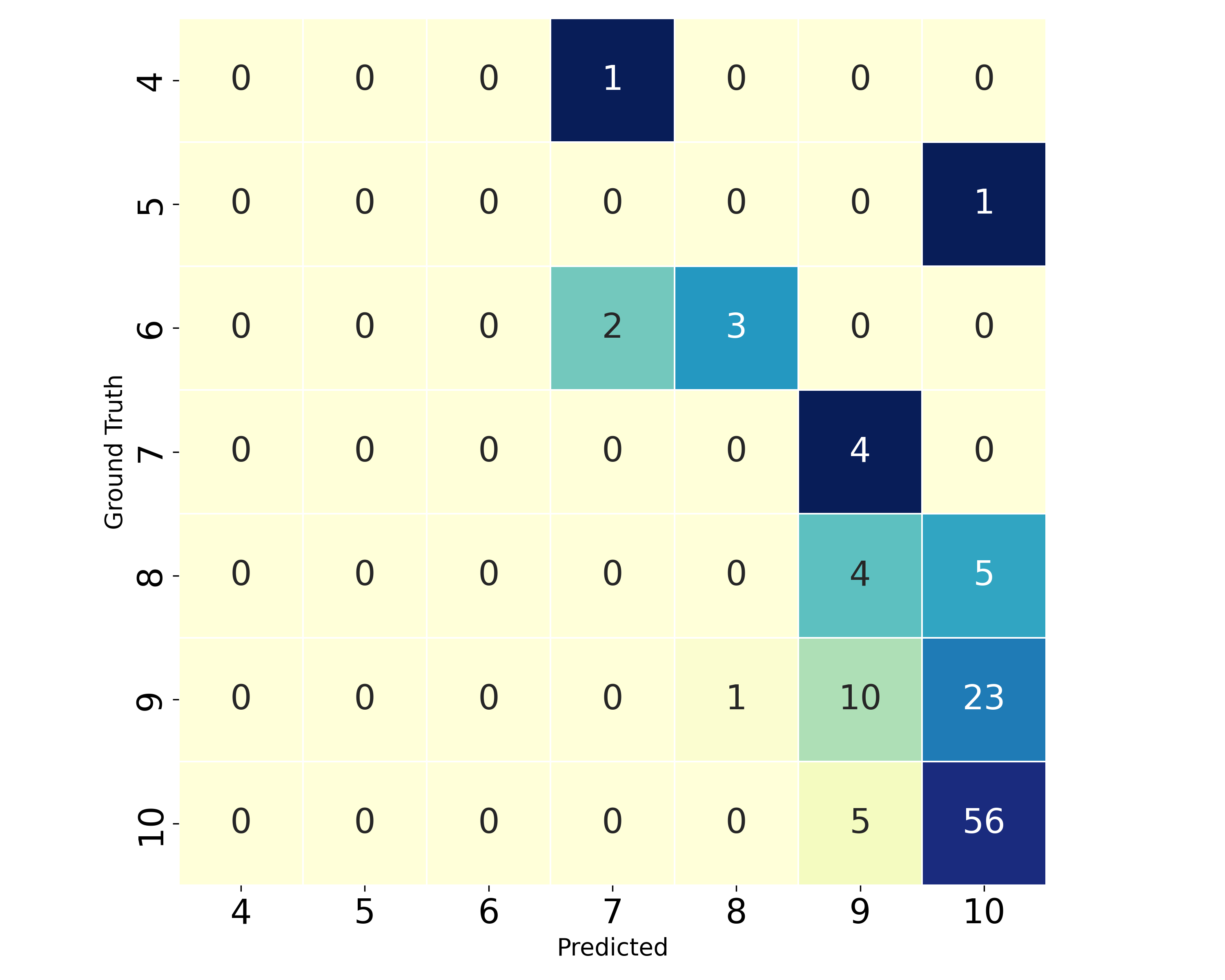}
    \caption{Word-For-Word}
    \label{fig:cm_wfw}
  \end{subfigure}
  \hfill
  \begin{subfigure}[b]{0.45\textwidth}
    \centering
    \includegraphics[width=0.9\linewidth]{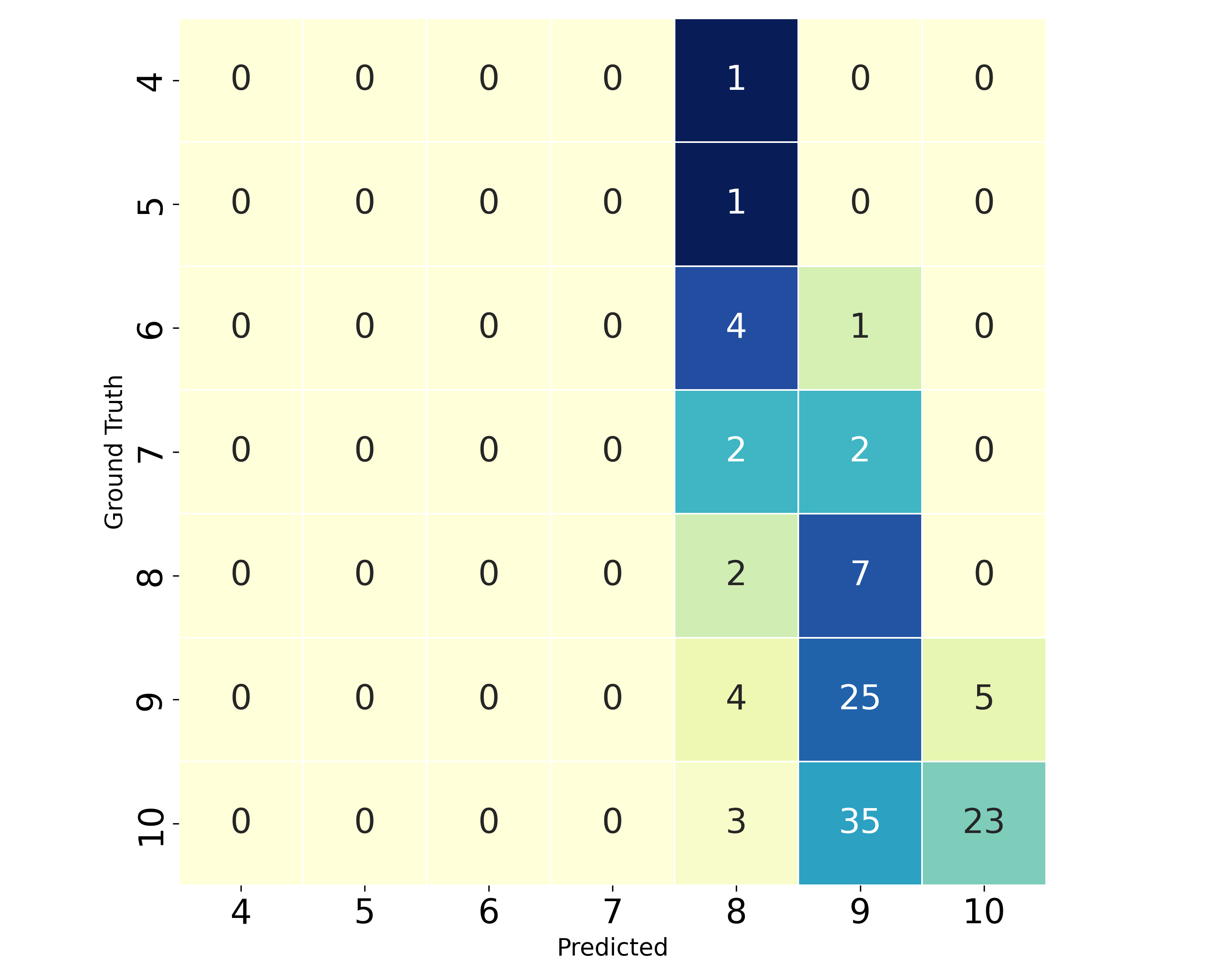}
    \caption{Llama 3.1}
    \label{fig:cm_llama}
  \end{subfigure}

  \vspace{0.5em} 

  \begin{subfigure}[b]{0.45\textwidth}
    \centering
    \includegraphics[width=0.9\linewidth]{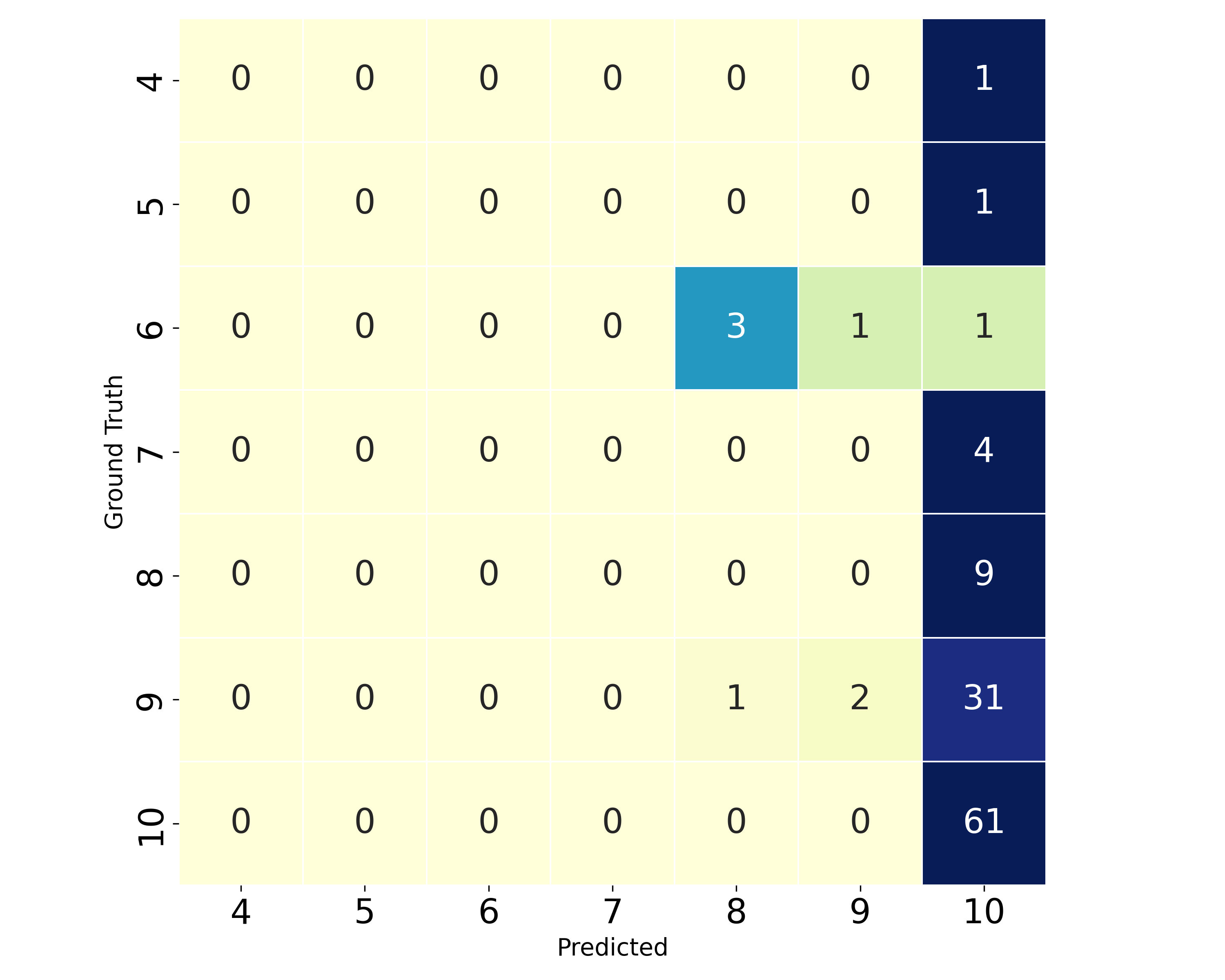}
    \caption{NER cosine}
    \label{fig:cm_ner}
  \end{subfigure}
  \hfill
  \begin{subfigure}[b]{0.45\textwidth}
    \centering
    \includegraphics[width=0.9\linewidth]{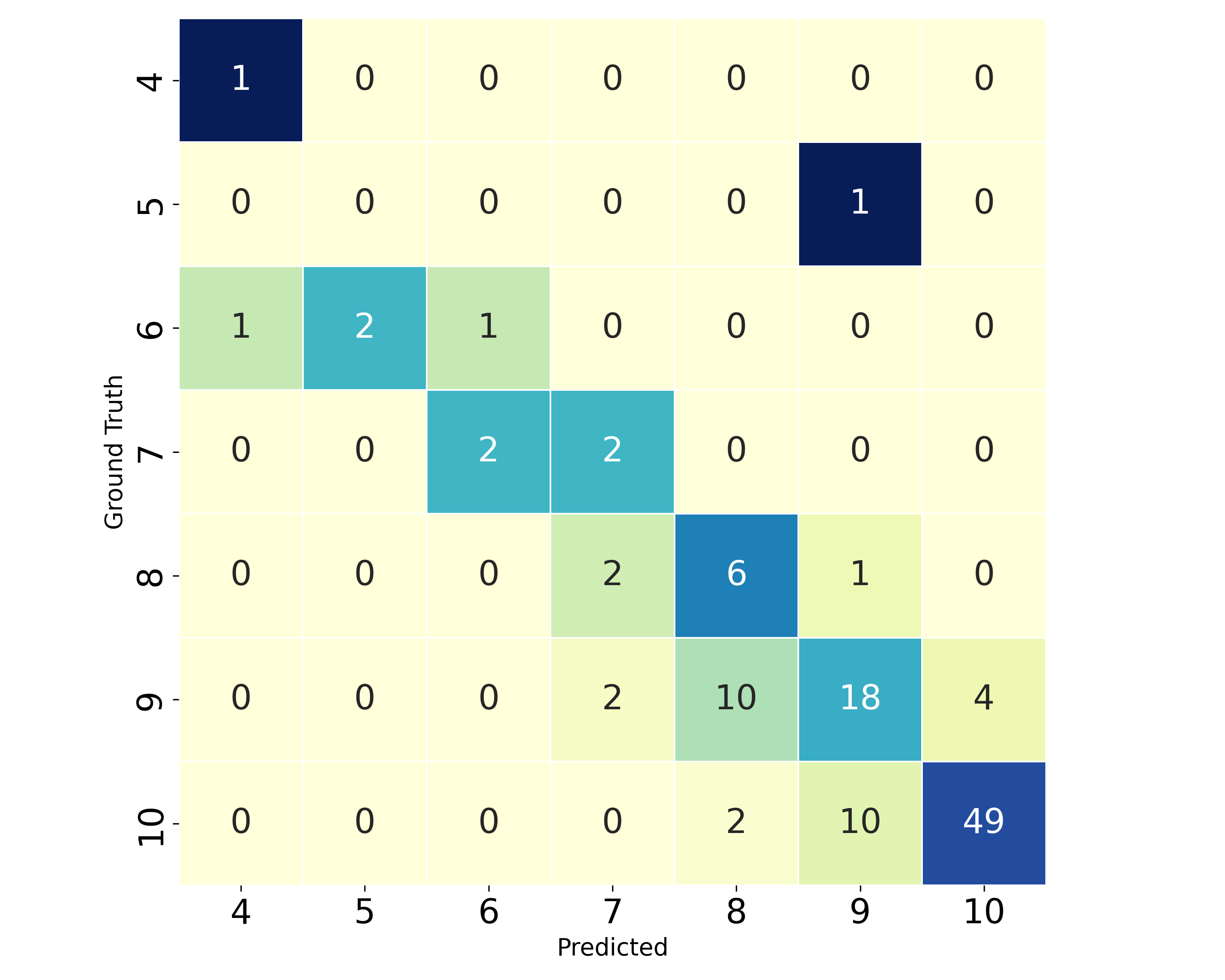}
    \caption{Llama-EntScore}
    \label{fig:cm_ours}
  \end{subfigure}

  \caption{Confusion matrices for predicted vs. ground truth scores. Llama-EntScore produces a tighter diagonal pattern, indicating more precise estimates.}
  \label{fig:conf_matrices}
\end{figure}

\begin{table}[htbp]
  \begin{center}
    \resizebox{.75\textwidth}{!}{
    \begin{tabular}{lccccc}
    \toprule
    \textbf{Method} & \textbf{Accuracy} & \textbf{Accuracy ($\pm 1$)} & \textbf{Precision} & \textbf{Recall} & \textbf{F1-Score} \\\midrule
    Word-For-Word & 0.57 & 0.88 & 0.16 & 0.17 & 0.16  \\
    Llama 3.1 & 0.43 & 0.90 & 0.19 & 0.19 & 0.16  \\
    Cosine (NER Only) & 0.55 & 0.83 & 0.18 & 0.15 & 0.12 \\
     \textbf{Llama-EntScore} & \textbf{0.67} & \textbf{0.94} & \textbf{0.38} & \textbf{0.46} & \textbf{0.40} \\
    \bottomrule
    \end{tabular}}
  \caption{Evaluation metrics for all methods. Scores are considered correct if they round to the same integer as the ground truth. $\pm 1$ accuracy measures predictions within one point.}
  \label{tab:results}
  \end{center}
\end{table}
Confusion matrices in Figure~\ref{fig:conf_matrices} illustrate the distribution of errors. While word-for-word and cosine scores skew rightward—suggesting consistent overestimation—Llama 3.1 exhibits vertical clustering near score 9. In contrast, Llama-EntScore forms a clean diagonal pattern, indicating more accurate predictions centred on true values.

\begin{figure}[h!]
    \centering
    \begin{subfigure}[b]{0.45\textwidth}
        \includegraphics[width=\textwidth]{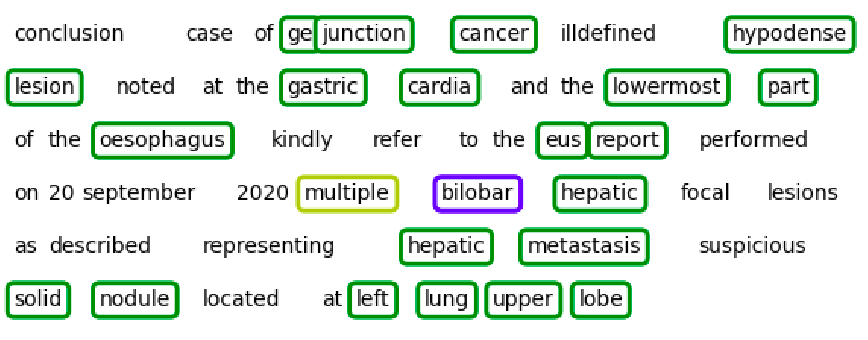}
        \caption{Preliminary report }
        \label{fig:entities_prelim}
    \end{subfigure}
    \hfill
    \begin{subfigure}[b]{0.45\textwidth}
        \includegraphics[width=\textwidth]{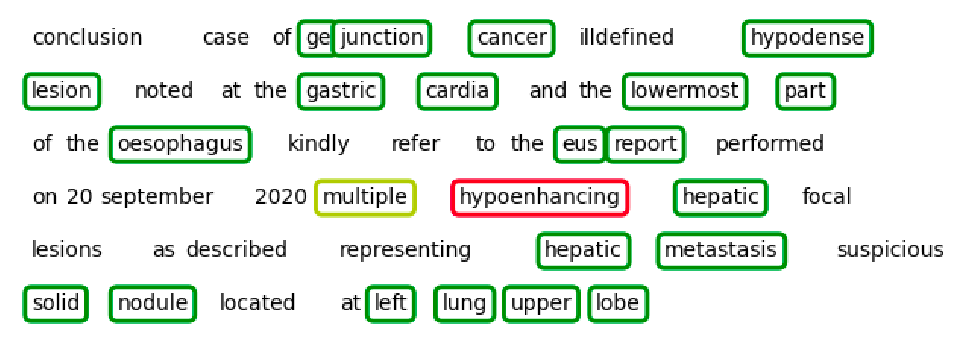}
        \caption{Final report}
        \label{fig:entities_final}
    \end{subfigure}
        \caption{Entity comparison visualisation. Colour codes: green = matched, yellow = mismatched, red = missing, blue = surplus.}
    \label{fig:entities}

\end{figure}

To enhance interpretability, we visualise the named entities found in each report, along with their classification as matched, mismatched, missing, or surplus (Figure~\ref{fig:entities}). This provides a concrete view of the differences driving the similarity score.

\begin{figure}[h!]
    \centering
    \includegraphics[width=\textwidth]{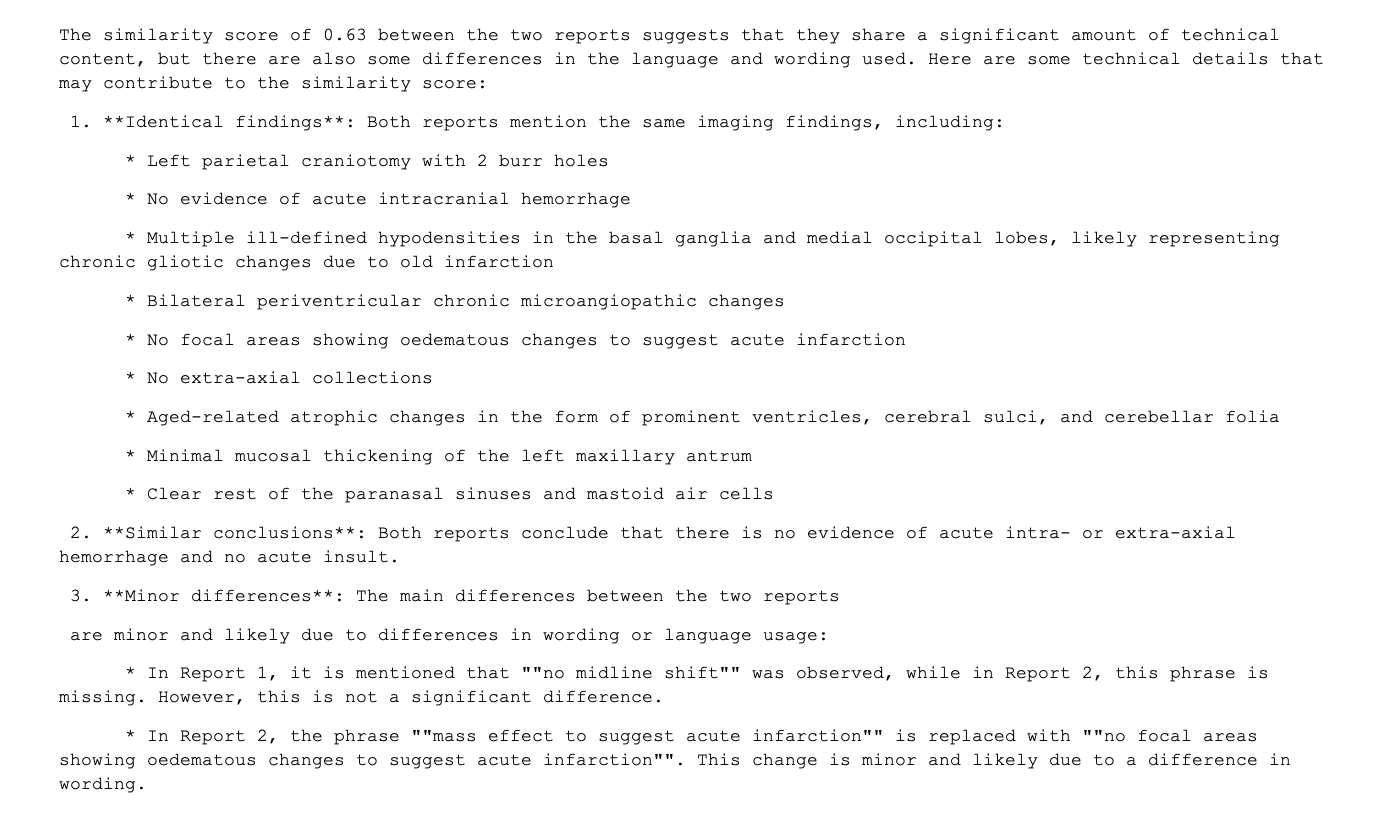} 
    \caption{Example LLM-generated explanation for a similarity score of 0.63 between a preliminary and final radiology report. The output highlights matched findings, similar conclusions, and minor wording differences.}
    \label{fig:llm_explanation}
\end{figure}

Finally, Figure ~\ref{fig:llm_explanation} includes an example explanation generated by the Llama-EntScore, grounded in the content of both reports and the similarity score. This narrative feedback highlights which findings were consistent, which differed, and why the assigned score is appropriate. These explanations serve as valuable educational feedback for junior radiologists reviewing their preliminary reports.

\section{Conclusion}
We present Llama-EntScore - a semantic similarity method for comparing final and preliminary radiology reports, combining NER with Llama 3.1. Our approach generates a quantitative similarity score for tracking progress and also gives an interpretation of the score that aims to offer radiologists valuable guidance in reviewing and refining their reporting. We find that Llama 3.1 delivers the most insightful output among the LLMs tested; however, its numerical scoring lacks the precision required to meet radiologists' needs. Llama-EntScore achieves the best alignment with radiologist similarity scores across all metrics. Additionally, our numerical score can be adjusted to weigh different types of discrepancies more or less heavily, enabling customisation to align with the preferences of different radiologist groups or different report data distributions. 
Despite its promising performance, Llama-EntScore has some limitations. It currently handles only four categories of differences, and its effectiveness depends on the accuracy of the NER component. Unrecognised medical terms may be overlooked, affecting comparison outcomes. Additionally, on average, the hybrid method takes 90 seconds to process a single report pair running on an NVIDIA GeForce RTX 2080 Ti, limiting its scalability for large-scale datasets.

Future work will focus on several directions: (1) Exploring how the choice of NER model affects overall performance for example by comparing different models on their accuracy with ground truth scores. Additionally, fine-tuning the NER model on domain-specific radiology corpora to improve performance on specialised terminology, (2) Expanding the categories of entity differences in reports to include distinctions such as severity, anatomical location errors, or omitted findings. Categories could also be merged with domain knowledge, enabling identification of differences specific to the type of scan. (3) Reducing computational overhead to enable near real-time processing for clinical integration. Additionally, we plan to explore interactive feedback mechanisms, allowing radiologists to refine the system’s outputs, and investigate its generalisability to other medical report types or specialties.

\begin{credits}
\subsubsection{\ackname} This work was supported by the Engineering and Physical Sciences Research Council (EPSRC) Impact Acceleration Account [grant number A100419], under the project AI Approach for Enhancing Radiology Reports.

\end{credits}

\bibliographystyle{ieeetr}
\bibliography{mybibliography}
\end{document}